\documentclass[10pt,twocolumn,letterpaper]{article}

\usepackage[pagenumbers]{cvpr} % To force page numbers, e.g. for an arXiv version

\usepackage{graphicx}
\usepackage{amsmath}
\usepackage{amssymb}
\usepackage{booktabs}

\usepackage{url}
\usepackage{epsfig}
\usepackage[accsupp]{axessibility}
\usepackage{helvet}   %Required
\usepackage{courier}  %Required
\usepackage{url}      %Required
\usepackage{xcolor}
\usepackage{multirow}
\usepackage{subcaption}
\usepackage{algorithm}
\usepackage{algpseudocode}
\usepackage[export]{adjustbox}
\usepackage{color}
\usepackage{pifont}
\definecolor{mycolor}{RGB}{147,112,219}
\definecolor{yccoloer}{RGB}{52, 235, 150}

\usepackage[pagebackref,breaklinks,colorlinks]{hyperref}

\usepackage[capitalize]{cleveref}
\crefname{section}{Sec.}{Secs.}
\Crefname{section}{Section}{Sections}
\Crefname{table}{Table}{Tables}
\crefname{table}{Tab.}{Tabs.}

\def\cvprPaperID{4889} % *** Enter the CVPR Paper ID here
\def\confName{CVPR}
\def\confYear{2022}
\begin{document}

%%%%%%%%% TITLE - PLEASE UPDATE
% \title{Adaptive Unbiased Teacher for Cross-Domain Object Detection}
% \title{Cross-Domain Object Detection via Adaptive Self-Training\vspace{-5mm}}
% \title{Adaptive Cross-Domain Object Detection via Self-Training\vspace{-5mm}}
% \title{Adaptive Teacher for Cross-Domain Object Detection\vspace{-5mm}}
\title{Cross-Domain Adaptive Teacher for Object Detection}

\author{Yu-Jhe Li$^{1}$\thanks{~Work done during the internship at Meta (Facebook).}\qquad Xiaoliang Dai$^{2}$\qquad Chih-Yao Ma$^{2}$\qquad Yen-Cheng Liu$^{3}$ \qquad Kan Chen$^{4}$\\ Bichen Wu$^{2}$\qquad Zijian He$^{2}$ \qquad Kris Kitani$^{1}$  \qquad Peter Vajda$^{2}$ \\
$^{1}$Carnegie Mellon University
\qquad
$^{2}$Meta (Facebook)
\qquad
$^{3}$Georgia Tech
\qquad
$^{4}$Waymo\\
{\tt\small \{\url{yujheli},\url{kkitani}\}\url{@cs.cmu.edu}, \{\url{xiaoliangdai},\url{cyma},\url{zijian},\url{wbc},\url{vajdap}\}\url{@fb.com},} \\
{\tt\small \url{ycliu@gatech.edu}, \url{kanchen@waymo.com}}
% \vspace{-7mm}
}
\maketitle
\begin{abstract}
% \vspace{-5mm}

We address the task of domain adaptation in object detection, where there is a domain gap between a domain with annotations (source) and a domain of interest without annotations (target). As an effective semi-supervised learning method, the teacher-student framework (a student model is supervised by the pseudo labels from a teacher model) has also yielded a large accuracy gain in cross-domain object detection. However, it suffers from the domain shift and generates many low-quality pseudo labels (\textit{e.g.,} false positives), which leads to sub-optimal performance. To mitigate this problem, we propose a teacher-student framework named Adaptive Teacher (AT) which leverages domain adversarial learning and weak-strong data augmentation to address the domain gap. Specifically, we employ feature-level adversarial training in the student model, allowing features derived from the source and target domains to share similar distributions. This process ensures the student model produces domain-invariant features.  Furthermore, we apply weak-strong augmentation and mutual learning between the teacher model (taking data from the target domain) and the student model (taking data from both domains). This enables the teacher model to learn the knowledge from the student model without being biased to the source domain. We show that AT demonstrates superiority over existing approaches and even Oracle (fully-supervised) models by a large margin. For example, we achieve 50.9\% (49.3\%) mAP on Foggy Cityscape (Clipart1K), which is 9.2\% (5.2\%) and 8.2\% (11.0\%) higher than previous state-of-the-art and Oracle, respectively. Code is released at \href{https://github.com/facebookresearch/adaptive_teacher}{the link}.

\end{abstract}
\section{Introduction}

% \begin{figure}[t!]
%   \centering
%   \begin{subfigure}{\linewidth}
%     {
%     \includegraphics[width=\linewidth]{TP_2.pdf}
%     }
%     \vspace{-7mm}
%     \caption{True positive ratio on pseudo labels}
%     \end{subfigure}
%     \begin{subfigure}{\linewidth}
%     {
%     \includegraphics[width=\linewidth]{FP_2.pdf}
%     }
%     \vspace{-7mm}
%     \caption{False positive ratio on pseudo labels}
%     \end{subfigure}
% %   \vspace{-5mm}
%   \caption{\textbf{The effectiveness of domain loss and weak-strong augmentation on pseudo labeling by Teacher model}. The figure shows the true positive and false positive ratio on the entire training set of Clipart1k (target) with PASCAL VOC as source. Due to inherent domain shift in the Teacher model, the Teacher model without domain loss generates noisy pseudo labels. The weak augmentation is able to stabilize pseudo labeling.}
%   \label{fig:teasor}
% %   \vspace{-5mm}
% \end{figure}

\begin{figure}[t!]
  \centering
  \includegraphics[width=\linewidth]{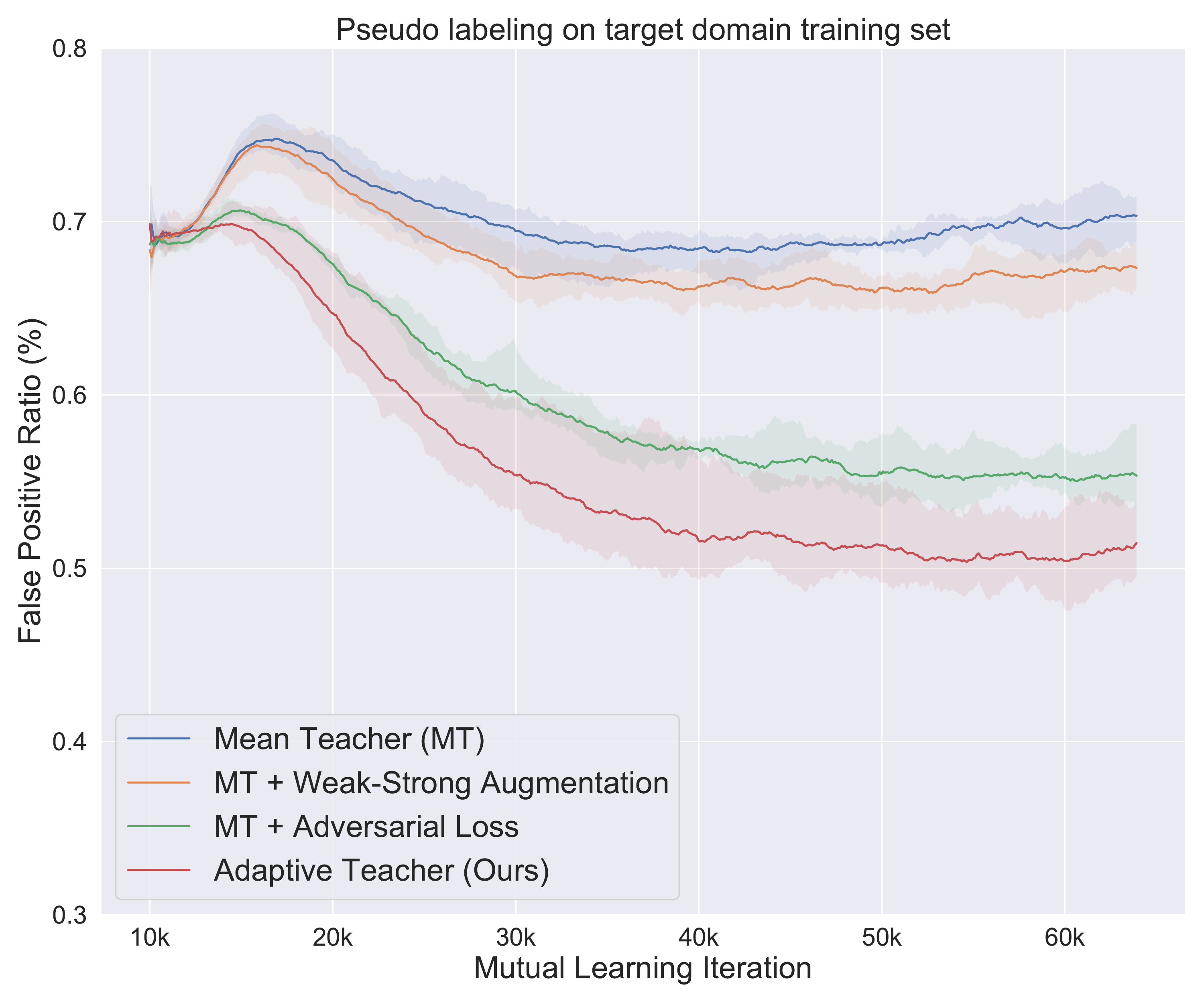}
  \vspace{-7mm}
  \caption{\textbf{The effectiveness of domain loss and weak-strong augmentation on pseudo labeling in Mean Teacher (MT)~\cite{tarvainen2017mean}}. The figure shows the false positive ratio on the training set of Clipart1k (target) with PASCAL VOC as source. We run 5 identical experiments for each setting and plot the error bound accordingly. Due to inherent domain shift in the Teacher model, it generates noisy pseudo labels without domain loss. The weak-strong augmentation is able to stabilize pseudo labeling.}
  \label{fig:teasor}
  \vspace{-5mm}
\end{figure}
% In the context of object detection, collecting annotations for target environment can be extremely expensive. 
% As the task of object detection is bringing pervasive impact on various real-world applications, collecting large-scale and diverse datasets with bounding box annotations still remains challenging since the labeling process is usually labor intensive.
Developing algorithms that can transfer the knowledge learned from one labeled dataset (\textit{i.e.,} source domain) to another unlabeled dataset (\textit{i.e.,} target domain) becomes increasingly important for object detection.
% if we have no annotations in a new environment (target domain) we need a method that can take an existing classifier (source domain) and adapt it to the new environment -- this is called cross-domain object detection.
% Hence, we address the task of cross-domain object detection where there is no bounding box annotations for the target domain, \textit{e.g.,} different weather and illumination conditions.
% Hence, developing models that can adapt to new environments without labeled data are highly desirable. 
Researchers have proposed various methods, such as domain classifier and adversarial learning~\cite{ganin2015unsupervised}, to address the task of cross-domain adaptation in object detection~\cite{chen2018domain,zhu2019adapting,saito2019strong,he2019multi,xu2020exploring,chen2020harmonizing,su2020adapting}. Even though these methods have led to accuracy improvement, solely using adversarial learning on the complex recognition task such as object detection is still limited. Hence, there is generally still a large performance gap from the Oracle model (fully supervision) on the target domain.
% the adversarial learning solely does not resolve the errors in bounding-box regressi with the 
%\kevin{Is this still true? Isn't Table 2 shows that all methods outperformed the Oracle? Xdai: added "generally". i guess these are the conventional approaches not the teacher-student settings in table 2}
% with solely domain classifier to address domain shift is quite limited without leveraging any self-supervised learning strategies on target domain.

To explore the potential of self-training on the unlabeled target domain for improved detection performance, researchers have exploited and extend the teacher-student self-training method from semi-supervised learning to domain adaptation~\cite{tarvainen2017mean}.
% To further improve the model's detection accuracy after domain adaptation, researchers have exploited and extend teacher-student self-training method from semi-supervised learning to domain adaptation~\cite{tarvainen2017mean}.
%which aim at similar scenarios where only a subset of the data has annotations.  
% These approaches typically involve a teacher model to generate pseudo labels thus to allow a student model to learn without annotations.
These approaches are able to learn without annotations by typically involving a teacher model to generate pseudo labels to update student model. These methods have led to notable accuracy gains in the domain adaptation scenario.
For example, MTOR~\cite{cai2019exploring} employs the Mean Teacher (MT)~\cite{tarvainen2017mean} as its pipeline to identify relations using region-level, inter-graph, and intra-graph consistency. 
Unbiased Mean Teacher (UMT)~\cite{deng2021unbiased} is proposed to augment the teacher-student framework with CycleGAN~\cite{zhu2017unpaired} and achieved further performance improvement.
%These methods extend the teacher-student method to domain adaptation and yield the current best adaptation accuracy.

%Quality of pseudo labeling
Despite the accuracy gain, the teacher-student framework still face a major challenge upon the settings of domain adaptation:  Unlike semi-supervised learning, the pseudo label generated from the teacher model usually contains a substantial amount of errors and false positives, as shown in Figure~\ref{fig:teasor}.  This is because the scenario of domain adaptation typically involves a large domain gap between the labeled data (source domain) and unlabeled data (target domain).  The teacher model is trained on, biased to, and only able to capture features precisely on the source domain, hence unable to provide high-quality pseudo labels in the target domain.  As a result, directly applying the teacher-student framework only leads to sub-optimal adaptation performance.

%However, the above methods still suffers the domain bias without any domain constraint (\textit{e.g.,} loss for aligning distribution across domains) on MT.
%Since annotations are only available on source data, the models can still be easily biased towards the source domain on both Teacher and Student model.

%To support the finding, we conduct experimental analysis in Figure~\ref{fig:teasor} to show how significant the domain loss (adversarial loss) is to generate reliable pseudo labels on target domain. We also discover that how weak augmentation (as opposed to applying complicated augmentation) proposed in Unbiased Teacher (\cite{liu2021unbiased}) is able to stabilize pseudo labeling.

% First, the above frameworks use strongly augmented (transformed or styled) images from source on Teacher model as perturbed samples which may prevent the Teacher model from generating reliable target-specific pseudo labels. Second, feeding more cross-domain augmented images in UMT without adaptation loss still suffers the domain bias in the Student model. Since annotations are only available on source data, the models can still be easily biased towards the source domain.
% Figure~\ref{fig:teasor} shows how Teacher model may suffer from bias toward without domain loss and generates noisy labels. 

To address this problem, we propose a self-training framework named Adaptive Teacher (AT) to mitigate the domain shift and improve the pseudo labeling quality on the target domain leveraging adversarial learning and mutual learning. Our model comprises of two separate modules: target-specific Teacher model and cross-domain Student model. %To ensure our Teacher model to generate precise and reliable target-specific pseudo labels following \cite{liu2021unbiased}, 
We also apply weak augmentation (only strong augmentation in Student model) and feed images from the target domain into the Teacher model, which we refer to as ``Weak-Strong augmentation", following Unbiased Teacher (UT)~\cite{liu2021unbiased}. This allows the teacher model to generate reliable pseudo labels without being affected by heavy augmentation. In addition, to mitigate the domain bias toward source domain in the Student model, we apply adversarial learning by introducing a discriminator with gradient reverse layer to align the distribution across two domains in the Student model.
% \yc{Adding citation for GRL?}
% This learning process is developed to bridge the domain bias toward source domain in the Student model.
With all the techniques, we observe the pseudo label quality improved significantly, as shown in Figure~\ref{fig:teasor}, where the false positive ratio is suppressed by up to 35\%.
% , where the true positive ratio is improved by up to 1.8$\times$ compared to the vanilla teacher-student framework. 
% That is, the false positive ratio is suppressed by up to 35\%.
This further leads to substantial accuracy gain across all the domain adaptation experiments and outperforms all existing methods. We summarize the contributions of this paper as follows:

% Our models are trained using two learning streams, which are mutual learning between Teacher and Student similar to Mean Teacher, and the adversarial learning in Student model. %
% Firstly, we follow the recent Teacher-Student learning framework from Unbiased Teacher~\cite{liu2021unbiased} that was originally proposed for semi-supervised object detection task, to generate pseudo-labels online for unlabeled images. 
% During the training, the Teacher generates pseudo-labels to train the Student, and the Student gradually updates the Teacher via Exponential Moving Average (EMA). 

\begin{itemize}
\itemsep -2pt
\item We demonstrate the limitation of the teacher-student framework in the domain adaptation scenario: The teacher model is biased toward the source domain and only able to produce low-quality pseudo labels on the target domain.
% Through the analysis of the pseudo labels on target, we discover the issue that current Mean Teacher generates substantial amount of false positives when having the bias toward source domain.
%
\item We propose a novel framework leveraging adversarial learning augmented mutual learning and weak-strong augmentation to address domain shift in cross-domain object detection.
%
% \item Our design of training pipeline contains target-specific Teacher model and cross-domain Student model, which resolves the bias toward source domain in both of Teacher and Student models with weak-strong augmentation.
%
\item Our method is able to deal with domain shift and outperform all existing SOTA by a large margin. For example, we achieve 50.9\% mAP on Foggy Cityscape, which is 9.2\% and 8.2\% higher than SOTA and Oracle (full supervision). 
\end{itemize}

% was initially proposed for semi- supervised learning. It consists of two models with identi- cal architecture, a student model and a teacher model. The student model is trained using the labeled data as standard, and the teacher model uses the exponential moving aver- age (EMA) weights of the student model. Each sample pre- diction of the teacher model can be seen as an ensemble of the student model’s current and earlier versions, therefore it is more robust and stable. By enforcing the consistency of teacher and student models using a distillation loss based on unlabeled samples, the student model is then guided to

% In this work, we are specifically interested in the cross-domain object detection problem, because the strong demands from real-world scenarios. For instance, in autonomous driving, robust object detection is needed in different weathers and lighting conditions. 
% Collecting annotations for all conditions can be extremely costly, and therefore models that can adapt to new environments without labeled data are highly desirable.

\section{Related Works}

\paragraph{Object Detection.}
Object detection is a task to localize the object and its location given an input image. Recently, deep models have shown to be effective in object detection with anchor-based approaches, \textit{e.g.}, Faster R-CNN~\cite{ren2015faster} which introduces Region Proposal Networks (RPN) to facilitate the proposal generation for region of interests (ROI). Afterwards, several anchor-based works~\cite{dai2016r,dai2017deformable,hu2018relation,lin2017feature,peng2018megdet,singh2018r} are proposed to improve the performance and efficacy. On the other hand, anchor-free methods are also proposed as one-stage detectors without the step of generating region proposals. YOLO~\cite{redmon2016you} produces bounding boxes and the confidence score jointly for multiple classes as regression task. Several of its variants~\cite{redmon2017yolo9000,redmon2018yolov3} are also proposed. SSD~\cite{liu2016ssd} is also built on top of YOLO yet leverages feature maps generated from different scales of images. For our work, we employ Faster R-CNN as the backbone for detection due to its flexibility.

\paragraph{Domain Adaptation.}
Given unlabeled data from target domain, unsupervised domain adaptation (UDA) or domain adaptation (DA) aims to learn a model from additional labeled source domain to achieve satisfactory performance on the target domain. Recently, it has demonstrated its effectiveness using deep neural networks. On one hand, some works have developed discrepancy-based methods that learn the representations by minimizing the domain discrepancy, which is also known as Maximum Mean Discrepancy (MMD)~\cite{long2015learning,long2017deep,long2016unsupervised}. Another line of domain adaptation is to map the domain distribution and treat it as a adversarial (min-max) optimization with a domain classifier~\cite{ganin2015unsupervised,ganin2016domain,tzeng2017adversarial,sankaranarayanan2018generate}. Some generative models such as CycleGAN~\cite{zhu2017unpaired} can also be seen as image-level domain adaptation.
% Self-ensembling~\cite{french2017self} advances Mean Teacher~\cite{tarvainen2017mean} for domain adaptation and achieves improved results on several cross-domain benchmarks related to recognition. 
% While these works above focus on common tasks of domain adaptation in recognition, recently several models also leverage attention mechanism in domain adaptation for other tasks, \textit{e.g.}, semantic segmentation~\cite{chen2018road,hoffman2016fcns,zhang2018fully}.
However, compared to these general vision tasks, the problem of object detection is more complicated since it has to predict the bounding box and class label for each object. Comparing with other recognition tasks, we aim at handling such challenging task of cross-domain object detection.

\begin{figure*}[t!]
  \centering
  \includegraphics[width=0.85\linewidth]{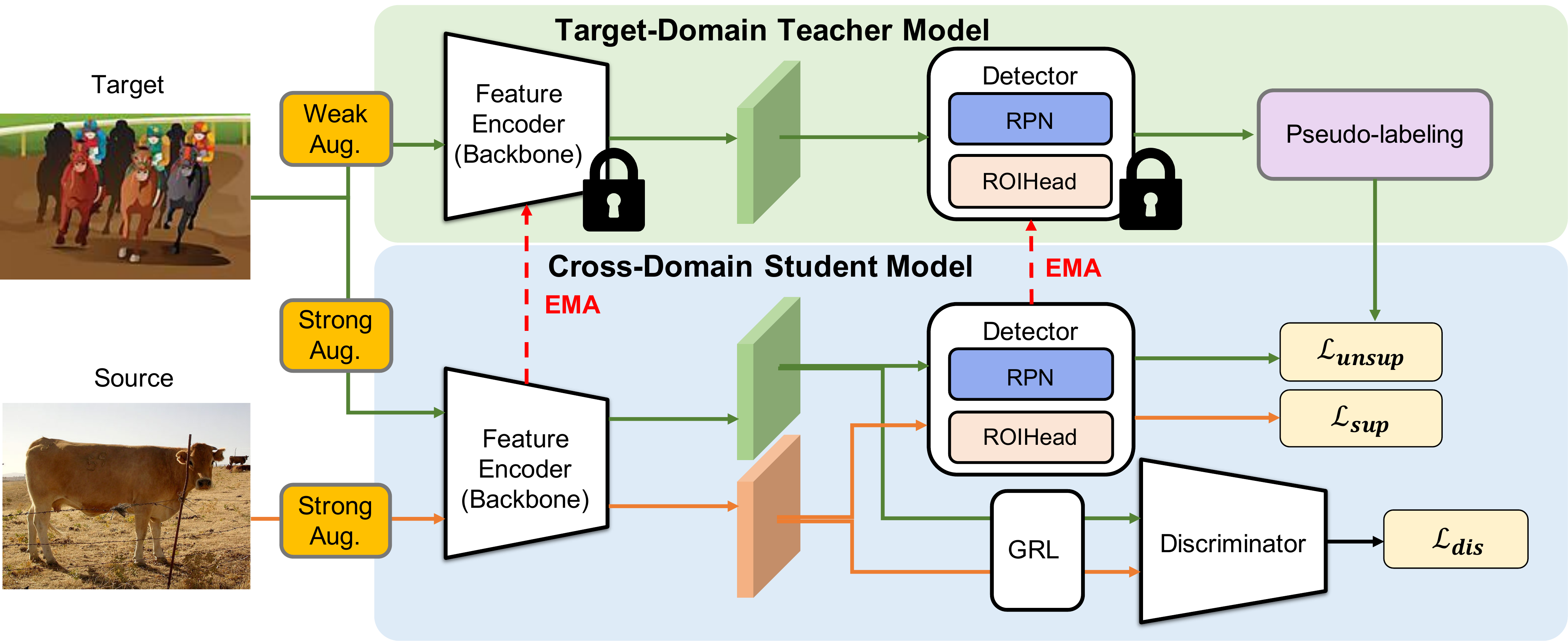}
  \vspace{-3mm}
  \caption{\textbf{Overview of our proposed Adaptive Teacher (AT).} Our model consists of two modules: 1) target-specific Teacher model for taking weakly-augmented images from target domain and 2) cross-domain Student model for taking strongly-augmented images from both domains. We train our model using two learning streams: Teacher-Student mutual learning and adversarial learning. The Teacher model generates pseudo-labels to train the Student while the Student updates the Teacher model with exponential moving average (EMA). The discriminator with gradient reverse layer is employed to align the distributions across two domains in Student model.}
  \label{fig:model}
%   \vspace{-5mm}
\end{figure*}

\paragraph{Cross-Domain Object Detection.}
Recently, more works pay more attention on domain adaptation in the task of object detection and propose various approaches. Some utilize adversarial learning with a gradient reverse layer (GRL) for mapping the feature across different domains in \cite{chen2018domain,zhu2019adapting,saito2019strong,he2019multi,xu2020exploring,chen2020harmonizing,su2020adapting}.
% Diversify \& Match (DM)~\cite{kim2019diversify} generates various distinctive domains from source domain and encourages features to be indistinguishable among the domains.
The annotation-level adaptation~\cite{khodabandeh2019robust,kim2019self,roychowdhury2019automatic} or curriculum learning~\cite{soviany2021curriculum} have also been proposed for the task of domain adaptation. Recently, another direction is to utilize Mean Teacher (MT)~\cite{tarvainen2017mean} which is originally proposed for semi-supervised learning on this task. MTOR~\cite{cai2019exploring} is proposed on top of MT and train its teacher network with enforcing the region-level, inter-graph, and intra-graph consistency. Similarly, Unbiased Mean Teacher (UMT)~\cite{deng2021unbiased} has been proposed to reduce the domain shift by augmenting the training samples with CycleGAN~\cite{zhu2017unpaired}. However, the above approaches are likely to suffer the same and inherent issue in Mean Teacher (MT), generating pseudo labels of low quality on target domain.
% In this work, we propose a framework based on MT and apply the adversarial learning strategies and weak-strong augmentation to address the domain shift in the task of cross-domain object detection.
\section{Adaptive Teacher}
\label{sec:method}

% \kevin{Overall comments on Sec.3: I think the majority of the section reads too similar to Unbiased Teacher. We spent quite a bit of space explaining the detail of the method, but they are parts that are identical from the Unbiased Teacher. This will likely give the reviewers impressions that we are basically just applying the Unbiased Teacher, and we don't have anything else to offer. 

% This is not the case. We can discuss the design decision we made or what analysis or things we found that finally led us to this final design choice.
% There are sections and paragraphs that I commented below, which I think we can add a little bit of story and discussion there to help set up the story of the paper/method.

% }

\subsection{Problem Formulation and Overview}
Before we demonstrate how our proposed method is able to mitigate the errors of pseudo labels in domain adaptation of object detection, we first review the problem formulation.
% For the problem of unsupervised domain adaptation in object detection, 
We are given $N_s$ labeled images $\mathcal{D}_s=\{(X_s, B_s, C_s)\}$ in source domain and $N_t$ unlabeled images $\mathcal{D}_t=\{X_t\}$ in target domain, where $B_s=\{b_s^i\}_{i=1}^{N_s}$ denotes the bounding box annotations and $C_s=\{c_s^i\}_{i=1}^{N_s}$ denotes corresponding class labels for source image $X_s=\{x_s^i\}_{i=1}^{N_s}$. There is no annotations for the target image $X_t=\{x_t^j\}_{j=1}^{N_t}$. The ultimate goal of cross-domain object detection is to design domain-invariant detectors by leveraging $\mathcal{D}_s$ and $\mathcal{D}_t$.

The overview of our framework is presented in Figure~\ref{fig:model}. Our AT framework consists of two modules: target-specific Teacher model and cross-domain Student model. The Teacher model only takes into the weakly-augmented images from target domain ($\mathcal{D}_t$) while the Student model takes strongly-augmented images from both domains ($\mathcal{D}_s$ and $\mathcal{D}_t$). We train our model using two training streams which are the Teacher-Student mutual learning and the adversarial learning strategies. To begin with, we train the object detector with the available source labeled data and initialize the feature encoder and detector. At the stage of mutual learning, we duplicate the initialized object detector into two identical detectors, \textit{i.e.,} Teacher and Student models. The Teacher generates pseudo labels to train the Student while the Student updates the knowledge it learned back to the Teacher via exponential moving average (EMA). Iteratively, the pseudo-labels for training the Student are improved. Furthermore, we employ the discriminator and the gradient reverse layer (GRL)~\cite{ganin2015unsupervised} for the adaptive learning (Sec.~\ref{sec:method:adaptive}) to align the distributions across two domains in Student model. This allows the Student model to reduce domain shifts and benefits the Teacher model to generate more accurate pseudo-labels.
% For the data augmentation, we apply random horizontal flip for weak augmentation and randomly add color jittering, grayscale, Gaussian blur, and cutout patches for strong augmentations.

\subsection{Mutual Learning between Teacher and Student}\label{sec:method:mutual}

%Based on Mean Teacher (MT)~\cite{tarvainen2017mean}, 
Following the teacher-student framework initially proposed for semi-supervised object detection, our model is also composed of two architecturally identical models: a Student model and a Teacher model. The Student model is learned by standard gradient updating, and the Teacher model is updated with the exponential moving average (EMA) of the weights from the student model. 
% Each sample prediction of 
% Since the Teacher model can be seen as an ensemble of the Student model’s current and earlier versions, its generated pseudo-labels are more robust and stable. 
To generate precise and accurate pseudo labels for target domain images, we feed the images with \textbf{weak augmentation} as input to the Teacher to provide reliable pseudo-labels while images with \textbf{strong augmentation} as inputs of the Student. Specifically, the target samples are augmented with randomly horizontal flipping and cropping as weak augmentation in Teacher model and randomly color jittering, grayscaling, Gaussian blurring, and cutting patches our for strong augmentations as perturbations.

\paragraph{Model Initialization.} Initialization is significant for the self-training framework since we rely on the Teacher to generate reliable pseudo-labels for target domain without annotations to optimize the Student model. To achieve this, we first use the available supervised source data $\mathcal{D}_s = \{({X}_s, {B}_s, {C}_s)\}$ to optimize our model with the supervised loss $\mathcal{L}_{sup}$. 
% \kevin{The following seems to be redundant unless later on we need to explicitly refer to these losses. Otherwise, these are considered to be common knowledge.}
% To employ Faster RCNN~\cite{ren2015faster} as the backbone of the detector, the supervised loss consists of four losses: the RPN classification loss $\mathcal{L}_{cls}^{rpn}$, the RPN regression loss $\mathcal{L}_{reg}^{rpn}$, the ROI classification loss $\mathcal{L}_{cls}^{roi}$, and the ROI regression loss $\mathcal{L}_{reg}^{roi}$.
Hence, the supervised loss for training and initializing the student model using the labeled source data can be defined as:
\begin{equation}
\begin{split}
\mathcal{L}_{sup}({X}_s, {B}_s, {C}_s)  = \mathcal{L}_{cls}^{rpn}({X}_s, {B}_s, {C}_s)+\mathcal{L}_{reg}^{rpn}({X}_s, {B}_s, {C}_s)\\
 +\mathcal{L}_{cls}^{roi}({X}_s, {B}_s, {C}_s)+\mathcal{L}_{reg}^{roi}({X}_s, {B}_s, {C}_s),
\end{split}
\end{equation}
where RPN loss $\mathcal{L}^{rpn}$ is the loss for learning the Region Proposal Network (RPN), which is designed to generate candidate proposals, and Region of Interest (ROI) loss $\mathcal{L}^{roi}$ is for the prediction branch of ROI. Both of RPN and ROI perform bounding box regression (reg) and classification (cls). We use binary cross-entropy loss for $\mathcal{L}_{cls}^{rpn}$ and $\mathcal{L}_{cls}^{roi}$, and $l_1$ loss for $\mathcal{L}_{reg}^{rpn}$ and $\mathcal{L}_{reg}^{roi}$.

\paragraph{Optimize Student with Target Pseudo-Labels}
% \kevin{The following paragraph is basically the same as how Unbiased Teacher did it. We don't have to spend this amount of space for this part (student is trained by the pseudo-labels), unless we are making modifications to it.}
% To address the lack of ground-truth labels in target domain, 
As the labels are not available in the target domain, we adopt the pseudo-labeling method to produce dummy labels on the images from target domain to train the Student. To filter out the noisy pseudo-labels, we set a confidence threshold $\delta$ on the predicted bounding boxes from the teacher model to remove the false positives. In addition, we exclude duplicated boxes prediction by non-maximum suppression (NMS) for each class. Hence, after obtaining the pseudo-labels from Teacher model on the images of target domain, we can update the Student with the loss as:
\begin{equation}
\begin{split}
\mathcal{L}_{unsup}({X}_t, \hat{{C}}_t)  = \mathcal{L}_{cls}^{rpn}({X}_t,  \hat{{C}}_t)
 +\mathcal{L}_{cls}^{roi}({X}_t, \hat{{C}}_t),
\end{split}
\end{equation}
where $\hat{\mathcal{C}}_t$ denotes the pseudo labels generated by the Teacher model on target domain. %Not
Here, unsupervised losses are not applied in the task of bounding box regression since the confidence score of predicted bounding boxes on the unlabeled data can only represent the confidence of the categories for each object instead of the locations for the produced bounding boxes.

% We would like to note that, since the incorrect bounding box regression may not be able to get filtered using confidence thresholding, we do not apply unsupervised losses for the task of bounding box regression following~\cite{liu2021unbiased}.

% random cropping, grayscale, Gaussian blur, and color jittering. The augmented target samples are only fed into student models as perturbations and then we take the instance predictions of teacher model to guide the student model. 

\paragraph{Temporally Update Teacher from Student.}
% By enforcing the consistency of teacher and student models using a distillation loss based on unlabeled samples, the student model is then guided to 
To obtain strong pseudo-labels from the target images following MT~\cite{tarvainen2017mean}, we apply Exponential Moving Average (EMA) to update Teacher model by temporally copying the weights of the student model. The update formula can be defined as:
\begin{equation}
\begin{split}
\theta_t \leftarrow \alpha \theta_t +(1-\alpha) \theta_s,
\end{split}
\end{equation}
where $\theta_t$ and $\theta_s$ denote the network parameters of Teacher and Student, respectively.

\subsection{Adversarial Learning to Bridge Domain Bias}\label{sec:method:adaptive}
% \kevin{Again, not sure about using "heal" here.}
% \kevin{Some context/story is missing here and made the following paragraph reads more like: we did A, B, and then C... We should add and discuss things that could make this paragraph more interesting and insightful, e.g., discussing why we decided to add the discriminator right after the feature encoder. Why not any where else?}
Since annotations are only available on source data, both of the Teacher and the Student can be easily biased towards the source domain during the mutual learning process. To be particular, the pseudo labels generated on target images from Teacher model are basically derived using the knowledge of the model trained with labels from source domain. As a result, we need to bridge the domain bias across source and target domains otherwise the Teacher model would generate noisy labels on target images and make the learning process collapse.  Thus, we introduce adversarial learning into the framework for aligning the distributions across two domains.  This leads to substantial false positive ratio reduction (20\% on MT+adversarial loss) in pseudo label generation, as shown in Figure~\ref{fig:teasor}. 

% \subsubsection{Domain adversarial learning}
Since Student model takes images from both domains, the adversarial loss is applicable on Student model to align two distribution.
To achieve adversarial learning, a domain discriminator $D$ is placed after the feature encoder $E$ (shown in Figure~\ref{fig:model}) on the Student model. The discriminator is aimed at discriminate where the derived feature $E(X)$ is from (source or target). Then we can define the probability of each input sample that belonging to the target domain as $D(E(X))$ and the probability of it belonging to source domain as $1 - D(E(X))$. We can update the domain discriminator $D$ using binary cross-entropy loss given the domain label $d$ for each of the input image. Specifically, images from the source domain are labeled as $d=0$ and images from target domain are labeled as $d=1$. The discriminator loss $\mathcal{L}_{dis}$ can be formulated as:
\begin{equation}
\begin{split}
\mathcal{L}_{dis} = - d \log D(E(X)) - (1-d) \log (1-D(E(X))) ,
\end{split}
\end{equation}

\begin{table*}[t]
%   \scriptsize
  \caption{The results of cross-domain object detection on the Clipart1k test set for \textbf{PASCAL VOC $\rightarrow$ Clipart1k} adaptation. The average precision (AP, in \%) on all classes is reported. The used backbone is ResNet-101 for fair comparison. We compare our method with SCL~\cite{shen2019scl}, SWDA~\cite{saito2019strong}, DM~\cite{kim2019diversify}, CRDA~\cite{xu2020exploring}, HTCN~\cite{chen2020harmonizing}, UMT~\cite{deng2021unbiased}, Source (F-RCNN), and Oracle (F-RCNN).}
  \vspace{-2.0mm}
  \centering
  \label{table:cli}
  \resizebox{\linewidth}{!}
  {
  \begin{tabular}{l|cccccccccccccccccccc|c}
  \toprule
  {Method}  & aero & bcycle & bird & boat & bottle &  bus & car & cat &  chair & cow & table & dog & hrs & m-bike & prsn & plnt & sheep &  sofa &  train &  tv & mAP\\
%   \multirow{3}{*}{Method} & \multirow{3}{*}{Source} & \multicolumn{3}{c|}{ConstructSite} & \multicolumn{3}{c|}{WILDTRACK} & \multicolumn{3}{c|}{Mars} & \multicolumn{3}{c}{DukeMTMC}\\
%   %
%   & \multicolumn{4}{c}{Test on: Div-Market}\\
%   \cmidrule{3-14} 
  %
%   \cmidrule{3-14} 
  %
%   & & Rank1 & Rank5 & mAP & Rank1 & Rank5 & mAP & Rank1 & Rank5 & mAP & Rank1 & Rank5 & mAP\\
  %
  \midrule
  Source & 23.0 & 39.6 & 20.1 & 23.6 & 25.7 & 42.6 & 25.2 & 0.9 & 41.2 & 25.6 & 23.7 & 11.2 & 28.2 & 49.5 & 45.2 & 46.9 & 9.1 & 22.3 &  38.9 & 31.5 & 28.8 {\color{red} (-16.2)}\\
%   Source (F-RCNNfpn) & 31.8 & 56.4 & 24.6 & 23.3 & 37.56 &  49.3 & 36.7 & 8.6 & 41.0 & 23.1 & 42.3 & 11.3 & 33.8 &  57.4 & 53.0 & 39.6 & 8.2 & 20.3 & 31.1 & 37.8 & 33.1\\
  \midrule
  SCL  & \textbf{44.7} & 50.0 & 33.6 & 27.4 & 42.2 & 55.6 & 38.3 & \textbf{19.2} & 37.9 & \textbf{69.0} & 30.1 & 26.3 & 34.4 &  67.3 & 61.0 & 47.9 & 21.4 & 26.3 & 50.1 & 47.3 & 41.5 {\color{red} (-3.5)}\\
  SWDA  & 26.2 &48.5 &32.6 &33.7 &38.5& 54.3 &37.1& 18.6& 34.8& 58.3 &17.0 &12.5 &33.8 &65.5 &61.6 &52.0 &9.3 & 24.9 &54.1 &49.1 &38.1 {\color{red} (-6.9)}\\
  DM & 25.8 & \textbf{63.2} &24.5 &42.4 &47.9& 43.1 &37.5 &9.1 &47.0 & 46.7 &26.8& 24.9& 48.1 & \textbf{78.7} & 63.0 &45.0 &21.3& 36.1 &52.3& 53.4 &41.8 {\color{red} (-3.2)}\\
  CRDA & 28.7 &55.3 &31.8 &26.0& 40.1& \textbf{63.6} &36.6 &9.4 &38.7 & 49.3 &17.6 &14.1 &33.3 &74.3 &61.3 &46.3 &22.3 &24.3 &49.1 &44.3 &38.3 {\color{red} (-6.7)}\\
  HTCN & 33.6 &58.9 &34.0 &23.4 &45.6 &57.0 &39.8 &12.0 &39.7 & 51.3 &21.1 &20.1 &39.1 &72.8 &63.0 &43.1 &19.3& 30.1& 50.2& 51.8 &40.3 {\color{red} (-4.7)}\\
  UMT & 39.6& 59.1& 32.4 &35.0 &45.1& 61.9&48.4 &7.5 &46.0 & 67.6 &21.4 & \textbf{29.5} &48.2 &75.9 &70.5 &56.7 & 25.9& 28.9 &39.4& 43.6 & 44.1 {\color{red} (-0.9)}\\
  \midrule
  AT & 33.8 & 60.9 & \textbf{38.6} & \textbf{49.4} & \textbf{52.4} & 53.9 & \textbf{56.7} & 7.5 & \textbf{52.8} & 63.5 & \textbf{34.0} & 25.0 & \textbf{62.2} & 72.1 & \textbf{77.2} & \textbf{57.7} & \textbf{27.2} & \textbf{52.0} & \textbf{55.7} & \textbf{54.1} & \textbf{49.3} {\color{blue} (+4.3)}\\
%   \midrule
%   AUTfpn w/o dis.  & 45.5 & 63.7 & 31.7 & 43.4 & 66.4 & 58.4 & 53.3 & 3.9 & 63.2 & 48.9 & 48.2 & 24.5 & 48.4 & 87.7 & 72.4 & 62.9 & 26.7 & 35.1 & 47.1 & 55.2 & 49.3\\
%   AUTfpn   & 39.8 & 65.5 & 34.9 & 42.0 & 63.8 & 66.0 & 54.7 & 6.3 & 62.1 & 45.3 & 44.6 & 27.1 & 53.9 & 89.0 & 73.9 & 65.4 & 29.1 & 29.0 & 45.6 & 54.1 & 51.2\\
%   AUTfpn (pyramid dis.) & 42.5 & 77.7 & 39.3 & 45.0 & 66.6 & 71.1 & 58.3 & 3.6 & 62.5 & 55.4 & 48.2 & 24.1 & 51.6 & 79.2 & 73.9 & 63.1 & 31.1 & 36.6 & 47.0 & 55.0 & \textbf{52.5}\\
  \midrule
%   Oracle (F-RCNN) & 13.2 & 38.8 & 41.5 & 33.2 & 20.9 & 57.9 & 52.8 &22.8 & 42.0 & 23.4 & 36.7 & 32.3 & 29.8 & 44.0 & 69.9 & 42.4 & 45.9 & 17.3 & 52.6 & 45.2 & 38.3\\
  Oracle & 33.3 & 47.6 & 43.1 & 38.0 & 24.5 & 82.0 & 57.4 & 22.9 & 48.4 & 49.2 & 37.9 & 46.4 & 41.1 & 54.0 & 73.7 & 39.5 & 36.7 & 19.1 & 53.2 & 52.9 & 45.0\\
%   Oracle (F-RCNNfpn) & 22.4 & 35.9 & 40.2 & 40.9 & 28.3 & 49.2 & 61.4 & 18.1 & 45.6 & 53.2 & 45.5 & 32.0 & 34.5 & 66.9 & 74.1 & 46.0 & 49.8 & 23.1 & 54.2 & 53.0 & 43.6\\

  \bottomrule
  \end{tabular}
  }
  \vspace{-3.5mm}
\end{table*}
On the other hand, the feature encoder $E$ is encouraged to produce features that confuse the discriminator $D$ while the discriminator $D$ aim to distinguish which domain the derived features are from. Hence, such adversarial optimization objective function can be defined as the following:
\begin{equation}
\begin{split}
\mathcal{L}_{adv} = \max_{E} \min_{D} \mathcal{L}_{dis}.
\end{split}
\end{equation}
% \kevin{Same as above. Some context/story is missing here and made the following paragraph reads more like: we did A, B, and then C... We can try to set up the story a bit here, e.g., talks about why do we choose to use GRL specifically? What do we expect from using GRL? Why didn't we propose a new version for it? Was it perfect? Can we improve it?}
Fortunately, to simply the min-max optimization, we can append an additional Gradient Reverse Layer (GRL)~\cite{ganin2015unsupervised} between the feature encoder and the discriminator to produce reverse gradient.
% This allows us to learn the domain-invariant feature for $E(X)$. 
During the gradient calculation, the GRL negates the gradients that pass back and the gradients of feature encoder $E$ is calculated in an opposite direction. This helps to maximize the discriminator loss for learning $E$ while we only need to minimize the objective $\mathcal{L}_{dis}$.
With the above discriminator loss, our Student model resolves the domain bias in visual features and helps Teacher to generate precise pseudo labels after several EMA updates.

We would like to note that, the design of adversarial learning in the Student model of our Adaptive Teacher is reasonable for two reasons. First, since we only feed images from target domain into Teacher model to avoid domain bias on Teacher model, the process of aligning two domains could be preferable in Student model which takes images across two domains. Feeding images from source domain like \cite{cai2019exploring,deng2021unbiased} may bring more bias toward source domain to both Teacher and Student models. Second, adversarial learning is a min-max learning problem and requires loss function to update the model. Since Student model is updated via objective losses, applying adversarial loss to the Student model is a simple and suitable way in the standard learning of the Mean Teacher.

\begin{table}[t]
%   \scriptsize
  \caption{The results of cross-domain object detection on the Watercolor2k test set for \textbf{PASCAL VOC $\rightarrow$ Watercolor2k} adaptation. The average precision (AP, in \%) on all classes is reported. The used backbone is ResNet-101 for fair comparison.}
\label{table:water}
  \vspace{-2.0mm}
  \centering
  \resizebox{\linewidth}{!}
  {
  \begin{tabular}{l|cccccc|c}
  \toprule
  {Method} & bicycle & bird & car & cat & dog & person& mAP\\
  \midrule
  Source & 84.2 & 44.5 & 53.0 & 24.9 & 18.8 & 56.3 & 46.9 {\color{red} (-3.7)}\\
  \midrule
  SCL~\cite{shen2019scl} & 82.2 & 55.1 & 51.8 & 39.6 & 38.4 & 64.0 & 55.2 {\color{blue} (+4.8)}\\
  SWDA~\cite{saito2019strong} & 82.3 & 55.9 & 46.5 & 32.7 & 35.5 & 66.7 & 53.3 {\color{blue} (+2.7)}\\
  UMT~\cite{deng2021unbiased} & 88.2 & 55.3 & 51.7 & \textbf{39.8} & \textbf{43.6} & 69.9 & 58.1 {\color{blue} (+7.5)}\\
  \midrule
  AT  & \textbf{93.6} & \textbf{56.1} & \textbf{58.9} & 37.3 & 39.6 & \textbf{73.8} & \textbf{59.9} {\color{blue} (+9.3)}\\
  \midrule
  Oracle & 51.8 & 49.7 & 42.5 & 38.7 & 52.1 & 68.6 & 50.6 \\
  \bottomrule
  \end{tabular}
  }
\end{table}
\subsection{Full Objective and Inference}
The total loss $\mathcal{L}$ for training our proposed AUT is summarized as follows:
\begin{equation}
  \begin{split}
  \mathcal{L} & = \mathcal{L}_\mathrm{sup} + \lambda_\mathrm{unsup}\cdot\mathcal{L}_\mathrm{unsup} + \lambda_\mathrm{dis}\cdot\mathcal{L}_\mathrm{dis}, \\
  \end{split}
  \label{eq:fullobj}
\end{equation}
where $\lambda_\mathrm{unsup}$ and $\lambda_\mathrm{dis}$ are the hyper-parameters used to control the weighting of the corresponding losses. We note that $\mathcal{L}_\mathrm{sup}$ and $\mathcal{L}_\mathrm{unsup}$ are developed to learn the feature encoder and detector in the Student model while $\mathcal{L}_\mathrm{dis}$ is introduced to update the feature encoder and discriminator. The Teacher model is only updated through EMA discussed in the Sec~\ref{sec:method:mutual}.

% Last but not the least, both the Teacher and the Student can jointly improve the accuracy of object detection on target domain during the learning interactions. This is achieved since the Teacher generates more and more reliable and stable pseudo-labels for target domain. On the other hand, in terms of EMA update the Teacher can be seen as the temporal ensemble of the cross-domain Student models with a momentum in different time steps. Such ensemble mechanism ensures the Teacher on target domain is consistently superior than the Student (noted in \cite{liu2021unbiased}). As the result, during the inference stage we only keep the target-specific Teacher model for evaluating on the target testing dataset. 

\section{Experiment}
\label{sec:EXP}

\begin{table*}[t]
%   \scriptsize
  \caption{The results and comparison on cross-domain object detection on the Foggy Cityscapes test set for \textbf{Cityscapes $\rightarrow$ Foggy Cityscapes} adaptation. The average precision (AP, \%) on all classes is presented. The used backbone is VGG-16 for fair comparison.}

  \vspace{-2.0mm}
  \centering
  \label{table:city}
  \resizebox{0.65\linewidth}{!}
  {
  \begin{tabular}{l|cccccccc|c}
  \toprule
  {Method}  & bus & bicycle & car & mcycle &person &rider& train &truck& mAP\\
%   \multirow{3}{*}{Method} & \multirow{3}{*}{Source} & \multicolumn{3}{c|}{ConstructSite} & \multicolumn{3}{c|}{WILDTRACK} & \multicolumn{3}{c|}{Mars} & \multicolumn{3}{c}{DukeMTMC}\\
%   %
%   & \multicolumn{4}{c}{Test on: Div-Market}\\
%   \cmidrule{3-14} 
  %
%   \cmidrule{3-14} 
  %
%   & & Rank1 & Rank5 & mAP & Rank1 & Rank5 & mAP & Rank1 & Rank5 & mAP & Rank1 & Rank5 & mAP\\
  %
  \midrule
  Source (F-RCNN) & 20.1 & 31.9 & 39.6 & 16.9 &  29.0 & 37.2 & 5.2 & 8.1 & 23.5 {\color{red} (-19.2)}\\
%   Source (F-RCNNfpn)& & 28.41 & 38.61 & 44.9 & 26.16 & 38.94 & 47.43 & 13.46 & 12.51 & 31.3 \\
  \midrule
  SCL~\cite{shen2019scl}  & 41.8 & 36.2 & 44.8 &33.6& 31.6& 44.0& 40.7& 30.4& 37.9 {\color{red} (-4.8)}\\
  DA-Faster~\cite{chen2018domain}  & 35.3 &27.1& 40.5& 20.0 &25.0 &31.0 &20.2& 22.1& 27.6 {\color{red} (-15.1)}\\
  SCDA~\cite{zhu2019adapting}  & 39.0 &33.6 &48.5 &28.0 &33.5 &38.0 &23.3& 26.5& 33.8 {\color{red} (-8.9)}\\
  SWDA~\cite{saito2019strong}  & 36.2 & 35.3 & 43.5 & 30.0 &  29.9 & 42.3 & 32.6 & 24.5 & 34.3 {\color{red} (-8.4)}\\
  DM~\cite{kim2019diversify}  & 38.4 &32.2 &44.3 &28.4 &30.8& 40.5 &34.5 & 27.2 & 34.6 {\color{red} (-8.1)}\\
  MTOR~\cite{cai2019exploring}  & 38.6  &35.6 & 44.0 & 28.3 & 30.6 & 41.4 & 40.6  &21.9 & 35.1  {\color{red} (-7.6)}\\
  MAF~\cite{he2019multi}  & 39.9 &33.9& 43.9& 29.2 &28.2& 39.5& 33.3& 23.8& 34.0 {\color{red} (-8.7)}\\
  iFAN~\cite{zhuang2020ifan} & 45.5& 33.0 &48.5 &22.8 &32.6 &40.0 &31.7& 27.9& 35.3 {\color{red} (-7.4)}\\
  CRDA~\cite{xu2020exploring} & 45.1 &34.6& 49.2 &30.3&32.9 &43.8& 36.4 &27.2 &37.4 {\color{red} (-5.3)}\\
  HTCN~\cite{chen2020harmonizing} & 47.4 &37.1& 47.9 &32.3& 33.2 &47.5 &40.9 &31.6& 39.8 {\color{red} (-2.9)}\\
  UMT~\cite{deng2021unbiased} & \textbf{56.5} & 37.3& 48.6 &30.4& 33.0& 46.7 &46.8& 34.1& 41.7 {\color{red} (-1.0)}\\
  \midrule
%   F-RCNN + dis. &  & 37.1 & 41.4 & 54.9 & 31.1 &  38.3 & 46.4 & 16.5 & 22.5 & 36.0\\
  
%   Baseline (F-RCNN + FPN) \\
%   F-RCNNfpn + dis.  &  & 35.83 & 44.76 & 53.34 & 23.55 &  44.58 & 51.58 & 23.14 & 17.44 & 36.78\\
%   F-RCNNfpn + pyramid dis.  &  & 36.11 & 45.9 & 53.43 & 29.92 & 46.25 & 53.74 & 11.65 & 19.15 & 37.02\\
%   \midrule
%   AUT$_\mathrm{base}$ w/o dis & & 58.56 & 53.57 & 63.97 & 38.53 & 45.46 & 55.32 & 43.75 & 33.64 & 49.07\\
%   AUT$_\mathrm{base}$ & & 52.82 & 50.09 & 62.88 & 35.33 & 43.93 & 53.77 & 38.01 &30.52& 45.92\\
%   AUTfpn$_\mathrm{base}$ w/o dis. & & 64.35 & 57.03 & 73.95 & 45.25 & 60.87 & 64.97 & 52.75 & 35.86 & 56.88 \\
%   AUTfpn$_\mathrm{base}$ & & 58.02 &57.57 &71.91 &43.87 & 58.96 & 61.79 & 44.41 &34.22& 53.84 \\
%   AUTfpn$_\mathrm{base}$ (pyramid dis.) & & \\
%   \midrule
%
%
% AUT w/o $\mathcal{L}_{unsup}$ \& EMA &  & 37.1 & 41.4 & 54.9 & 31.1 &  38.3 & 46.4 & 16.5 & 22.5 & 36.0\\
%   AUT w/o Aug. & & 52.8 & 50.1 & 62.9 & 35.3 & 43.9 & 53.8 & 38.0 &30.5& 45.9\\
%   AUT w/o $\mathcal{L}_{dis}$  &  & \textbf{57.5} & \textbf{52.6} & 63.8 & 37.5 & 45.2 & \textbf{55.3} & 46.8 & 31.8 & 49.3\\
%   MT & 52.82 & 50.09 & 62.88 & 35.33 & 43.93 & 53.77 & 38.01 &30.52& 45.92\\
%   MT & 52.82 & 50.09 & 62.88 & 35.33 & 43.93 & 53.77 & 38.01 &30.52& 45.92\\
  AT  &  56.3 & \textbf{51.9} & \textbf{64.2} & \textbf{38.5} &  \textbf{45.5} & \textbf{55.1} & \textbf{54.3} & \textbf{35.0} & \textbf{50.9} {\color{blue} (+8.2)}\\
%   \midrule
%   AUTfpn w/o dis. &  & 65.48 & 58.45 & 74.53 & 46.32 & 60.56 & 64.05& 47.16 & 37.41 & 56.7 {\color{blue} (+15)}\\
%   AUTfpn  &  & 63.6 & 55.94 & 73.97 & 46.37 & 60.2 & 63.48 & 56.69 & 38.97 & 57.4 {\color{blue} (+15.7)}\\
%   AUTfpn (pyramid dis.)  &  & 64.97 & 59.82 & 74.65 & 48.37 & 60.77 & 64.91 & 56.48 & 39.93 & \textbf{58.8} {\color{blue} (+17.1)}\\
  \midrule
  Oracle (F-RCNN) &  50.3 & 40.7 & 61.3 & 32.5 & 43.1 & 49.8 & 35.1 & 28.6 & 42.7 \\
%   Oracle (F-RCNNfpn) &  & 62.37 & 48.19 & 71.05 & 38.26 & 56.65 & 61.07 & 47.21 & 32.74 & 52.19 \\

  \bottomrule
  \end{tabular}
  }
  \vspace{-3.5mm}
\end{table*}

\subsection{Datasets}
We conduct our experiments on five public datasets, including Cityscapes~\cite{cordts2016cityscapes}), Foggy Cityscapes~\cite{sakaridis2018semantic}, PASCAL VOC ~\cite{everingham2010pascal}, Clipart1k~\cite{inoue2018cross}, and Watercolor2k~\cite{inoue2018cross}.

\textbf{Cityscapes.}
Cityscapes~\cite{cordts2016cityscapes} is collected by capturing images from outdoor street scenes in normal weather conditions from 50 cities, which has diverse scenes. It contains 2,975 images for training and 500 images for validation with dense pixel-level labels. The annotations of bounding boxes are converted from instance segmentation labels.

\textbf{Foggy Cityscapes.} Foggy Cityscapes~\cite{sakaridis2018semantic} is synthesized from the images in the Cityscapes. Therefore, it has the same train/test split as Cityscapes. It simulates the condition of foggy weather according to depth information provided in Cityscapes and generates three levels of foggy weather.

\textbf{PASCAL VOC.} PASCAL VOC~\cite{everingham2010pascal} contains 20 categories of common objects from real world with bounding box and class annotations. Following~\cite{saito2019strong,shen2019scl}, the dataset is combined from PASCAL VOC 2007 and 2012 with total 16,551 images.

\textbf{Clipart1k.} Clipart1k~\cite{inoue2018cross} contains clipart images and shares the same 20 classes with PASCAL VOC. Yet, it exhibits a large domain shift from PASCAL VOC. We follow the practice in \cite{saito2019strong,shen2019scl} and split it into training and test sets, containing 500 images each.

\textbf{Watercolor2k.} Watercolor2k~\cite{inoue2018cross} contains watercolor style images, which consists of images from 6 classes and shares with the same classes in PASCAL VOC dataset. Following \cite{saito2019strong,shen2019scl}, the dataset is splitted halfly into training set and testing sets and each contains 1000 images.

% \begin{figure}[t!]
%   \centering
%   \includegraphics[width=\linewidth]{fig/Learning curve.pdf}
%   \caption{\textbf{Mutual Learning curve on Clipart1k dataset}. Increasing weights of $\lambda_{dis}$ can achieve improved performance and stable learning curve.}
%   \label{fig:curve}
% %   \vspace{-5mm}
% \end{figure}

\subsection{Implementation Details}
Following \cite{chen2018domain} and \cite{saito2019strong}, we employ Faster RCNN~\cite{ren2015faster} as the base detection model in our Adaptive Teachera and implement it using Detectron2. Either of the network ResNet-101~\cite{he2016deep} or VGG16~\cite{simonyan2014very} pre-trained on ImageNet~\cite{deng2009imagenet} is used as the backbone according to the settings. Following the implementation of Faster RCNN with ROI-alignment~\cite{he2017mask}, we scale all images by resizing the shorter side of the image to $600$ while maintaining the image ratios. For the hyperparameter, we set the $\lambda_{unsup}=1.0$ and $\lambda_{dis}=0.1$ for all the experiments for simplicity. We set the confidence threshold as $\delta=0.8$. For the initialization stage of training the framework described in Sec.~\ref{sec:method:mutual}, we train the AT using the source labels for 10k iterations. Then we copy the weights to both Teacher and Student models in the beginning of mutual learning and train the AT for 50k iterations. We set the learning rate as 0.04 during the entire training stage without applying any learning rate decay.
We optimize the network using Stochastic Gradient Descent (SGD).
% with weight decay of 0.0005 and momentum of 0.9.
The used data augmentation methods include random horizontal flip for weak augmentation , and randomly color jittering, grayscaling, Gaussian blurring, and cuting out patches are utilized for strong augmentations. The weight smooth coefficient parameter of the exponential moving average (EMA) for the teacher model is set to 0.9996. Each experiment is conducted on 8 Nvidia GPU V100 with the batch size of 16 and implemented in PyTorch.

\begin{table*}[t]
%   \scriptsize
  \caption{The results of domain generalization on unseen target dataset, which leverages the labeled source data and another domain without supervision. The average precision (AP, \%) is reported. The backbone is ResNet-101 for fair comparison. ``WS Aug." indicates weak-strong augmentation.}
\label{table:general}
  \vspace{-2.0mm}
  \centering
  \resizebox{\linewidth}{!}
  {
  \begin{tabular}{l|cccccc|c|cccccc|c}
  \toprule
  \multirow{2}{*}{Method} & \multicolumn{7}{c|}{\textbf{PASCAL VOC (sup.) \& Watercolor2k (unsup.) $\rightarrow$ Clipart1k}} &
  \multicolumn{7}{c}{\textbf{PASCAL VOC (sup.) \& Clipart1k (unsup.) $\rightarrow$ Watercolor2k}}  \\
  \cmidrule{2-15} 
   & bicycle & bird & car & cat & dog & person & mAP& bicycle & bird & car & cat & dog & person& mAP\\
  \midrule
  
  AT  & 78.6 & 30.1 & 40.3 & 10.9 & 32.6  & 72.8 & 44.5 {\color{blue} (+1.1)} & 91.2 & 55.2 & 60.4 & 37.0 & 39.6 & 69.8 & 56.7 {\color{blue} (-6.1)}\\
%   AUT w/o $\mathcal{L}_{dis}$ & 68.2 & 25.6 & 35.2 & 2.9 & 25.5 & 64.5 & 37.4 {\color{red} (-6.0)} & 82.1 & 49.0 & 55.6 & 29.5 & 25.4 & 66.2 & 50.4 {\color{red} (-0.2)}\\
%   AUT w/o WS Aug. & 73.2 & 29.7 & 38.8 & 9.0 & 28.6 & 69.2 & 41.3 {\color{red} (-2.1)} & 84.3 & 51.2 & 58.7 & 34.2 & 24.3 & 62.4 & 51.1 {\color{blue} (+0.5)} \\
%   AUT w/o $\mathcal{L}_{unsup}$ \& EMA & 54.8 & 23.4 & 31.6 & 3.1 & 15.0 & 51.4 & 30.9 {\color{red} (-12.5)} & 80.5 & 43.4 & 53.0 & 27.6 & 19.5 & 55.6 & 47.6 ({\color{red} (-3.0)}) \\
  MT~\cite{tarvainen2017mean} + WS Aug. & 68.2 & 25.6 & 35.2 & 2.9 & 25.5 & 64.5 & 37.4 {\color{red} (-6.0)} & 82.1 & 49.0 & 55.6 & 29.5 & 25.4 & 66.2 & 50.4 {\color{red} (-0.2)}\\
  MT~\cite{tarvainen2017mean} + $\mathcal{L}_{dis}$ & 73.2 & 29.7 & 38.8 & 9.0 & 28.6 & 69.2 & 41.3 {\color{red} (-2.1)} & 84.3 & 51.2 & 58.7 & 34.2 & 24.3 & 62.4 & 51.1 {\color{blue} (+0.5)} \\
  MT~\cite{tarvainen2017mean} & 64.8 & 23.4 & 34.6 & 3.1 & 22.0 & 61.4 & 34.2 {\color{red} (-9.2)} & 80.5 & 43.4 & 53.0 & 27.6 & 19.5 & 55.6 & 47.6 ({\color{red} -3.0}) \\
  \midrule
  Oracle & 47.6 & 39.1 & 51.4 & 20.1 & 38.4 & 69.7 & 43.4 & 51.8 & 49.7 & 42.5 & 38.7 & 52.1 & 68.6 & 50.6 \\
  \bottomrule
  \end{tabular}
  }
  \vspace{-3mm}
\end{table*}

\subsection{Experimental Settings and Evaluation}
We report the average precision (AP) of each class as well as the mean AP over all classes for object detection following existing works~\cite{chen2018domain,saito2019strong} for all of the experimental settings, which are described as follows:
% \kevin{the following experimental settings seem mostly follow what existing works are doing. If so, we should try to reduce the space used here. Try to make every sentence in the paper count :)}

\paragraph{Real to Artistic Adaptation.}
To begin with, we would like to benchmark the effectiveness of our model for addressing the large domain gap. In this setting, we test our model with the effect of domain shift between the real images and the artistic images. We use Pascal VOC as the source dataset and the Clipart1k or Watercolor2k as the target dataset. The backbone of ResNet-101~\cite{he2016deep} is used following the settings in existing works.

\paragraph{Adverse Weather Adaptation.}
% In real-world scenarios, object detectors may be applied under different weather conditions, especially for anonymous driving. 
For this setting, we evaluate our model on the domain shift between the image in normal weather and the image with adverse weather (foggy). The data from Cityscapes dataset is served as the source domain while the data from Foggy Cityscapes dataset is served as the target domain. The model is trained with the labeled images from Cityscapes and unlabeled images from Foggy Cityscapes in the experiments. The testing results on the validation set of Foggy Cityscapes are reported. Even though there are one-to-one mappings of images between Cityscapes and Foggy Cityscapes datasets, we do not use such information in all of the experiments. The backbone of VGG16~\cite{simonyan2014very} is used following previous settings.

\subsection{Results and Comparisons}
In this section, we report the performance of our Adaptive Teacher and other state-of-the-art approaches in Table~\ref{table:cli} and Table~\ref{table:city}. We additionally report the source-only model denoted ``Source (F-RCNN)'' as training the base Faster RCNN model with only source images as the lower bound benchmark. On the other hand, we also include an oracle model denoted ``Oracle (F-RCNN)'' as training a base Faster RCNN model using the images from target domain and ground truth annotations, which can be viewed as the upper bound benchmark. 

\paragraph{Real to Artistic Adaptation.}
The results of the setting: real to artistic adaptation on Clipart1k is presented in Table~\ref{table:cli} and the one on Watercolor2k is presented in Table~\ref{table:water}. We compare our method with several state-of-the-art approaches and report the performance gap between the oracle model (fully supervision) and each of the competitors.
% : SCL~\cite{shen2019scl}, SWDA~\cite{saito2019strong}, DM~\cite{kim2019diversify}, CRDA~\cite{xu2020exploring}, HTCN~\cite{chen2020harmonizing}, and UMT~\cite{deng2021unbiased}. 
% For the popular benchmark: clipark1k, some phenomenons can be observed in two folds. 
We observed that, first, our model achieves state-of-the-art performance at $49.3\%$ mAP and outperforms the recent competitor UMT by $5.2\%$ and other methods by a large margin. We note that, UMT using Mean Teacher already had significant performance improvement with augmented-styled training images. Yet, due to the inherent issue with the quality of pseudo labels in Mean Teacher on target domain, their model may also suffer large domain shift between real and artistic images when generate pseudo labels. On the other hand, our model mitigates the domain gap and achieve largely improved performance. Second, our model is the only one exceeding the oracle model on Clipart1k dataset, showing that the mutual learning adopted form Mean Teacher plus adversarial learning is capable to bridge the domain gap. 
% The margin around $11\%$ at mAP between our model and the Oracle model can be credited to the efficiency of leveraging both the source large dataset (PASCAL VOC) and the target dataset.
Similar observations can be found on experiments conducted on Watercolor2k.

\begin{figure*}[t!]
  \centering
  \includegraphics[width=0.9\linewidth]{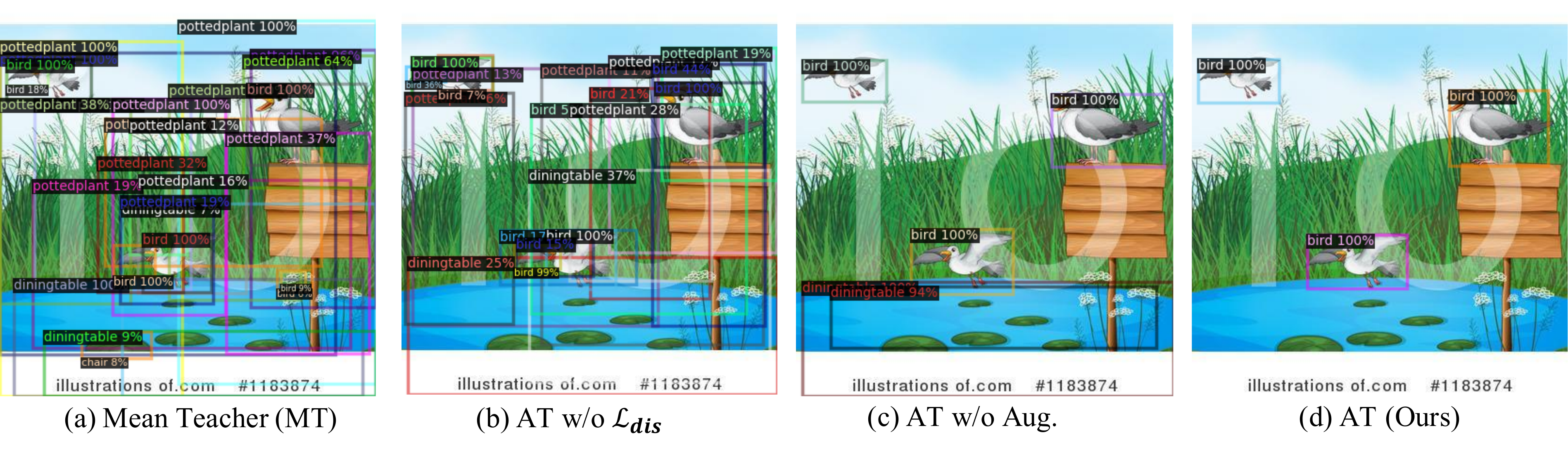}
  \vspace{-4mm}
  \caption{\textbf{Qualitative ablation studies on pseudo labels generated on the image from the training set of Clipart1k}. This figure show the importance of adversarial loss $\mathcal{L}_{dis}$ and weak-strong augmentation on pseudo labeling. Note that the thresholding is not applied here.}
  \label{fig:exp_quali_loss}
  \vspace{-4mm}
\end{figure*}

% \begin{figure*}[t!]
%   \centering
%   \includegraphics[width=0.95\linewidth]{fig/qual_compare.pdf}
%   \caption{\textbf{Qualitative results on Foggy Cityscapes and Clipart1k}. We compare our method with source only and oracle. The first two rows and the last two row are the results of Foggy Cityscapes and Clipart1k, respectively.}
%   \label{fig:exp_quali}
% %   \vspace{-5mm}
% \end{figure*}

\paragraph{Adverse Weather Adaptation.}
The results of the setting: normal weather to adverse weather adaptation is presented in Table~\ref{table:city}. 
% We compare our method with several state-of-the-art approaches containing  SCL~\cite{shen2019scl}, DA-Faster~\cite{chen2018domain}, SCDA~\cite{zhu2019adapting}, MTOR~\cite{cai2019exploring}, MAF~\cite{he2019multi}, iFAN~\cite{zhuang2020ifan}, SWDA~\cite{saito2019strong}, DM~\cite{kim2019diversify}, CRDA~\cite{xu2020exploring}, HTCN~\cite{chen2020harmonizing}, and UMT~\cite{deng2021unbiased}. Some phenomenons can also be observed in this table. 
We also report the performance gap between the oracle model (fully supervision) and each of the competitors.
When comparing to the state of the arts, we can see that,
first, our model also outperforms all of the state-of-the-art approaches by a large margin (more than 9\%). Among these methods, MTOR~\cite{cai2019exploring} and UMT~\cite{deng2021unbiased} are the two methods adopting Mean Teacher in their model. However, due to the problems discussed earlier regarding the augmentation in Teacher model and bias to source domain, both of their model suffer from generating noisy labels and lead to performance gap between our AT. Second, the performance of our model is able to exceed the oracle model by a large margin, which indicates that training the images with only the annotations from clear weather (high visibility) are sufficient to have satisfactory performance of the object detection on the adverse foggy weather (low visibility).

% \input{exp/Ablation_loss}
% \begin{figure*}[t!]
%   \centering
%   \includegraphics[width=\linewidth]{fig/abl_qual.pdf}
%   \caption{\textbf{Qualitative ablation studies on pseudo labels generated on the image from the training set of Clipart1k}. This figure show the importance of adversarial loss $\mathcal{L}_{dis}$ and weak-strong augmentation on pseudo labeling.}
%   \label{fig:exp_quali_loss}
% %   \vspace{-5mm}
% \end{figure*}
\begin{table}[t]
%   \scriptsize
  \caption{
%   \kevin{Let's try to not use "Source" and "Target" for each column. I think they can be a dedicated two-rows merged with "Method" cell?}
  The ablation studies on our AT. We report mean average precision (mAP, \%) on each of the experimental settings while ``WS Aug." indicates weak-strong augmentation.}

  \vspace{-2.0mm}
  \centering
  \label{table:abl}
  \resizebox{\linewidth}{!}
  {
  \begin{tabular}{l|c|c|cccc|c}
  \toprule
  \begin{tabular}[c]{@{}l@{}}Source: \\Target: \end{tabular} & \begin{tabular}[c]{@{}l@{}} PASCAL VOC\\Clipart1k\end{tabular} & \begin{tabular}[c]{@{}l@{}}PASCAL VOC\\Watercolor2k\end{tabular} & \begin{tabular}[c]{@{}l@{}}Cityscapes\\Foggy Cityscapes \end{tabular}\\
%   \multirow{3}{*}{Method} & \multirow{3}{*}{Source} & \multicolumn{3}{c|}{ConstructSite} & \multicolumn{3}{c|}{WILDTRACK} & \multicolumn{3}{c|}{Mars} & \multicolumn{3}{c}{DukeMTMC}\\
%   %
%   & \multicolumn{4}{c}{Test on: Div-Market}\\
%   \cmidrule{3-14} 
  %
%   \cmidrule{3-14} 
  %
%   & & Rank1 & Rank5 & mAP & Rank1 & Rank5 & mAP & Rank1 & Rank5 & mAP & Rank1 & Rank5 & mAP\\
  %
  \midrule
  AT  & 49.3 & 59.9 & 50.9  \\
  AT w/o $\mathcal{L}_{dis}$  & 40.6 {\color{red} (-8.7)} & 55.5 {\color{red} (-4.4)} & 48.7 {\color{red} (-2.2)} \\
  AT w/o WS Aug. & 45.3 {\color{red} (-4.0)} & 55.1 {\color{red} (-4.8)} & 45.9 {\color{red} (-5.0)}  \\
  AT w/o $\mathcal{L}_{unsup}$ \& EMA & 31.6 {\color{red} (-17.7)} & 50.2 {\color{red} (-9.7)} & 36.0 {\color{red} (-14.9)} \\

  \bottomrule
  \end{tabular}
  }
%   \vspace{-3.5mm}
\end{table}

\begin{figure}[h]
  \centering
  \includegraphics[width=\linewidth]{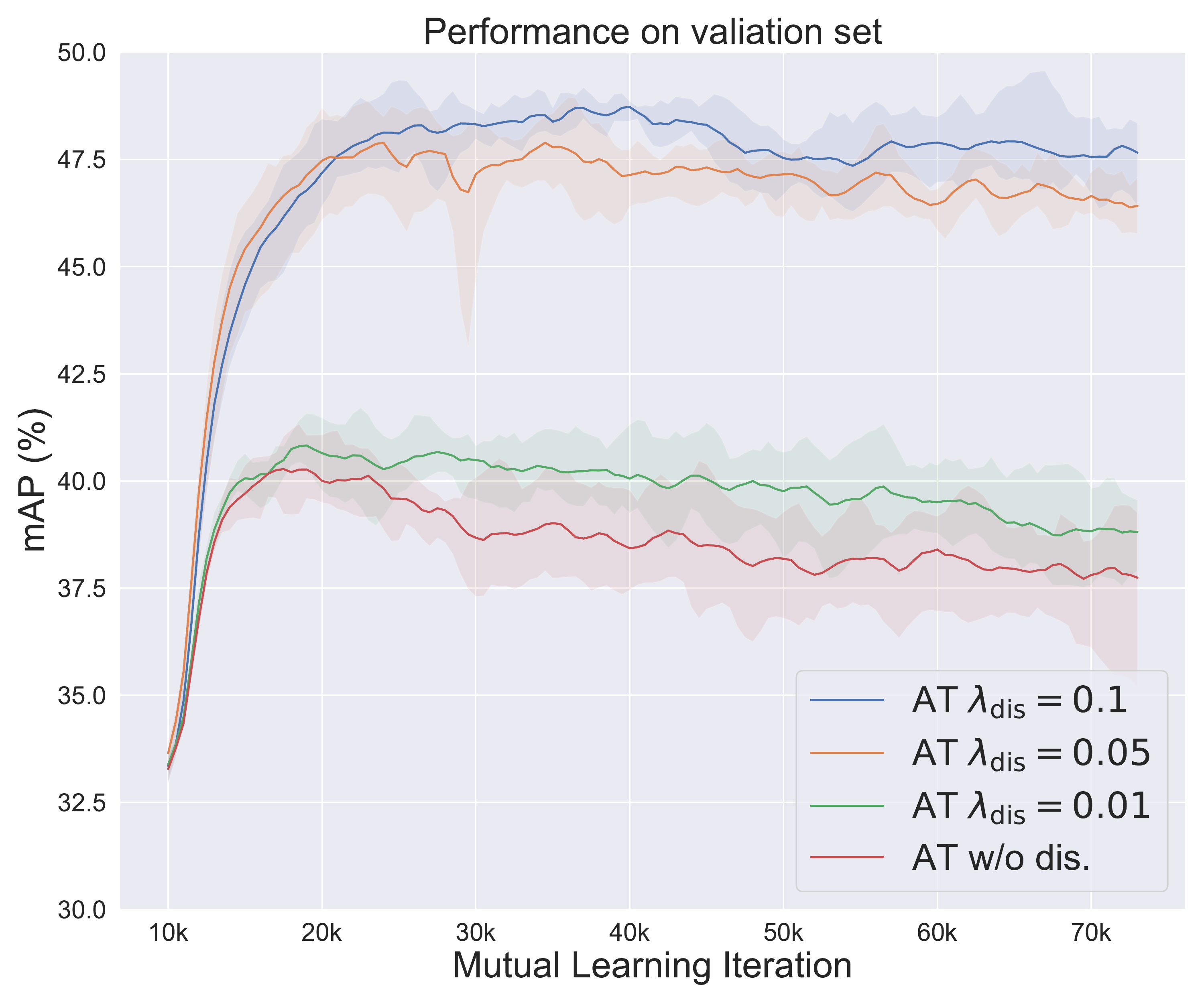}
  \vspace{-7mm}
  \caption{\textbf{Mutual Learning curve on Clipart1k dataset}. We run 5 identical experiments for each settings and plot the error bound accordingly in the figure. The results show that increasing weights of $\lambda_{dis}$ can achieve improved performance and stable learning curve. }
  \label{fig:curve}
  \vspace{-3mm}
\end{figure}

\subsection{Experiments on domain generalization}
As we observe that our AT outperforms all of the Oracle models on the three benchmark domain adaptation datasets, we are more interested in the ability of generalization of our model on the unseen domains. We define such problem as domain generalization: Instead of focusing on the model's accuracy on the target domain, we further generalize the model to a completely unseen domain and evaluate it's generalization capability.  In this section, we further conduct two experimental settings and compare our AT with the baseline model MT~\cite{tarvainen2017mean}: 
\begin{itemize}
\item Train: {PASCAL VOC (supervised) \& Watercolor2k (unsupervised) $\rightarrow$ Test: Clipart1k}
\item {Train: PASCAL VOC (supervised) \& Clipart1k (unsupervised) $\rightarrow$ Test: Watercolor2k}
\end{itemize}
In each of the setting, we train the model on the source real dataset with labels and another artistic dataset without labels. We then inference the model on the target dataset which is unseen during the training. We only train on the overlapped classes (6 classes) between the Clipart1k and Watercolor2k, and presented the results in the Table~\ref{table:general}. From the table, we can observe that our model achieves superior performance comparing with the Oracle model and MT. This shows that our model is able to generalize to unseen domain without observing any target images. In addition, each of the ablation studies on either adversarial loss or augmentation on MT also show the importance of their roles in our proposed AT.

\subsection{Ablation studies}
We further conduct ablation studies on each of important components in Table~\ref{table:abl} and also present the qualitative studies of pseudo labels in Figure~\ref{fig:exp_quali_loss}.
% The ablation study of loss weight of $\lambda_{dis}$ is also presented in Figure~\ref{fig:curve}.

\paragraph{Adversarial loss $\mathcal{L}_{dis}$.}
To further analyze the importance of adversarial learning in our Adaptive Teacher, we exclude the loss $\mathcal{L}_{dis}$ in discriminator and report the performance on three experimental settings in Table~\ref{table:abl}. It can be observed that the 8.7\% and 4.4\% performance drop appears on Clipart1k and Watercolor2k in the scenario with larger domain gap (real to artistic adaptation). Yet, in another scenario with smaller domain gap (weather adaptation), only 2.2\% performance drop is observed. We can also observe that $\mathcal{L}_{dis}$ is able to largely reduce the ratio of false positives in pseudo labels generated by the Teacher model in Figure~\ref{fig:exp_quali_loss}.
On the other hand, we also analyze the weight $\lambda_{dis}$ of the adversarial loss $\mathcal{L}_{dis}$ in in Figure~\ref{fig:curve}. Some phenomenons can be observed in this figure in two folds. 
We can see that, first, increasing weights can lead to improved performance, which supports the effectiveness of the discriminator in our model. Second, without applying the adversarial loss, the performance of the model keeps dropping due to the error propagation coming from the noisy pseudo labels.
% Second, we see that while such hyperparameter $\lambda_{dis}$ need to be determined in advance, the results were not sensitive to their choices (either 0.05 or 0.1).\kevin{hmmm... I actually think this is sensitive ... I would suggest to move this to the appendix and just say that we did this experiment (without saying that it's not sensitive).} 
% In other words, with a sufficient weight of discriminator loss, the model will be able to have satisfactory performance. Third, the performance with the adversarial loss $\mathcal{L}_{dis}$ shows to be stable without decreasing drastically as the one without the loss.

\paragraph{Augmentation pipeline.}
We also benchmarked the effectiveness of weak-strong (WS) augmentation in our Adaptive Teacher, and around 4\% to 5\% performance drop is observed when it is excluded (Table~\ref{table:abl}). This demonstrates that the simple modification on the training pipeline (weak and strong augmentation for Teacher and Student, respectively) is vital. We can also observe that such augmentation pipeline is able to reduce the ratio of false positives in pseudo labels generated by the Teacher model in Figure~\ref{fig:exp_quali_loss}.

\paragraph{$\mathcal{L}_{unsup}$ \& EMA.}
Similarly, we ablated the importance of utilizing Mean Teacher in Table~\ref{table:abl} as previous works (\textit{i.e.,} excluding the mutual learning and the Teacher model from our model) and report the performance of the Student model for cross-domain training with only strong augmentation and adversarial loss $\mathcal{L}_{dis}$. We can see that there is a significant performance drop, thus the performance gain mainly came from the mutual learning with pseudo labels on target domain.

% Even with only adversarial loss $\mathcal{L}_{dis}$, our model only improves the source model: Source (F-RCNN) by a small margin on Clipart1k in Table~\ref{table:cli}.\kevin{I couldn't find the experimental result for the last sentence.}

% \input{exp/clipart_exp}
% \input{exp/city_exp}
% \begin{figure*}[t!]
%   \centering
%   \includegraphics[width=\linewidth]{fig/quali.pdf}
%   \caption{\textbf{Qualitative results on Foggy Cityscapes and Clipart1k}. We compare our method with source only and oracle. The first two rows and the third row are the results of Foggy Cityscapes and Clipart1k, respectively.}
%   \label{fig:exp_quali}
% %   \vspace{-5mm}
% \end{figure*}

% \input{exp/clipart_more}
% \input{exp/city_more}

\section{Conclusion}
In this paper, we proposed an novel framework to address the task of cross-domain object detection. With the introduced target-domain Teacher model and cross-domain Student model, the framework is able to generate correct pseudo labels on the target domain via mutual learning. Our design of training pipeline with proper augmentation strategies and adversarial learning also resolve the bias toward source domain in both Teacher and Student model. The experiments on two benchmarks confirmed the effectiveness and superiority of our model for cross-domain object detection. The extensive experiments of ablation studies also demonstrated our proposed model trained without seeing both labels nor images on target domain  outperform the Oracle model which is trained with fully supervision.

\noindent\textbf{Acknowledgement: }We thank Meta (Facebook) for the sponsorship and the computing resources.

%%%%%%%%% REFERENCES
{\small
\bibliographystyle{ieee_fullname}
\bibliography{egbib}
}

\end{document}